\documentclass[11pt,a4paper]{article}
\usepackage[hyperref]{emnlp2020}
\usepackage{times}
\usepackage{latexsym}

\usepackage{graphicx}
\usepackage{lipsum}  
\usepackage{booktabs}
\usepackage{adjustbox}
\usepackage{url}
\usepackage{amssymb}
\usepackage{amsmath}
\usepackage{multirow}
\usepackage{tabularx}
\usepackage[normalem]{ulem}

\usepackage{microtype}

\aclfinalcopy % Uncomment this line for the final submission
\usepackage{color}

\usepackage[normalem]{ulem}

\title{{W}hy do you think that? {E}xploring {F}aithful {S}entence--{L}evel {R}ationales {W}ithout {S}upervision}

\author{Max Glockner \and Ivan Habernal \and Iryna Gurevych \\
  Ubiquitous Knowledge Processing Lab (UKP-TUDA) \\
  Department of Computer Science, Technische Universität Darmstadt \\
  \url{https://www.ukp.tu-darmstadt.de}\\
  }

\date{}

\begin{document}
\maketitle
\begin{abstract}

Evaluating the trustworthiness of a model's prediction is essential for differentiating between  `right for the right reasons' and `right for the wrong reasons'. 
Identifying textual spans that determine the target label, known as faithful \emph{rationales},  usually relies on pipeline approaches or reinforcement learning. However, such methods either require supervision and thus costly annotation of the rationales or employ non-differentiable models. We propose a differentiable training--framework to create models which output faithful rationales on a sentence level, by solely applying supervision on the target task. 
To achieve this, our model solves the task based on each rationale individually and learns to assign high scores to those which solved the task best. Our evaluation on three different datasets shows competitive results compared to a standard BERT blackbox while exceeding a pipeline counterpart's performance in two cases. We further exploit the transparent decision--making process of these models to prefer selecting the correct rationales by applying direct supervision, thereby boosting the performance on the rationale--level.\footnote{Code available at \url{https://github.com/UKPLab/emnlp2020-faithful-rationales}}
\end{abstract}
\section{Introduction}
Large pre-trained language models, such as BERT~\citep{devlin2018bert} or RoBERTa~\citep{liu2019roberta} gain impressive results on a large variety of NLP tasks, including reasoning and inference \citep{Rogers.et.al.2020.BERT}.
Despite this success, research shows that their strong performance can rely, to some extent, on dataset--specific artifacts and not necessarily on the ability to solve the underlying task \citep{gururangan-etal-2018-annotation, schuster-etal-2019-towards, gardner2020evaluating}. 
Thus, these observations undermine the models' trustworthiness and impede their deployment in situations where `blindly trusting' the model is deemed irresponsible \cite{Sokol2020}.
\begin{figure}[t!]
    \includegraphics[width=\linewidth]{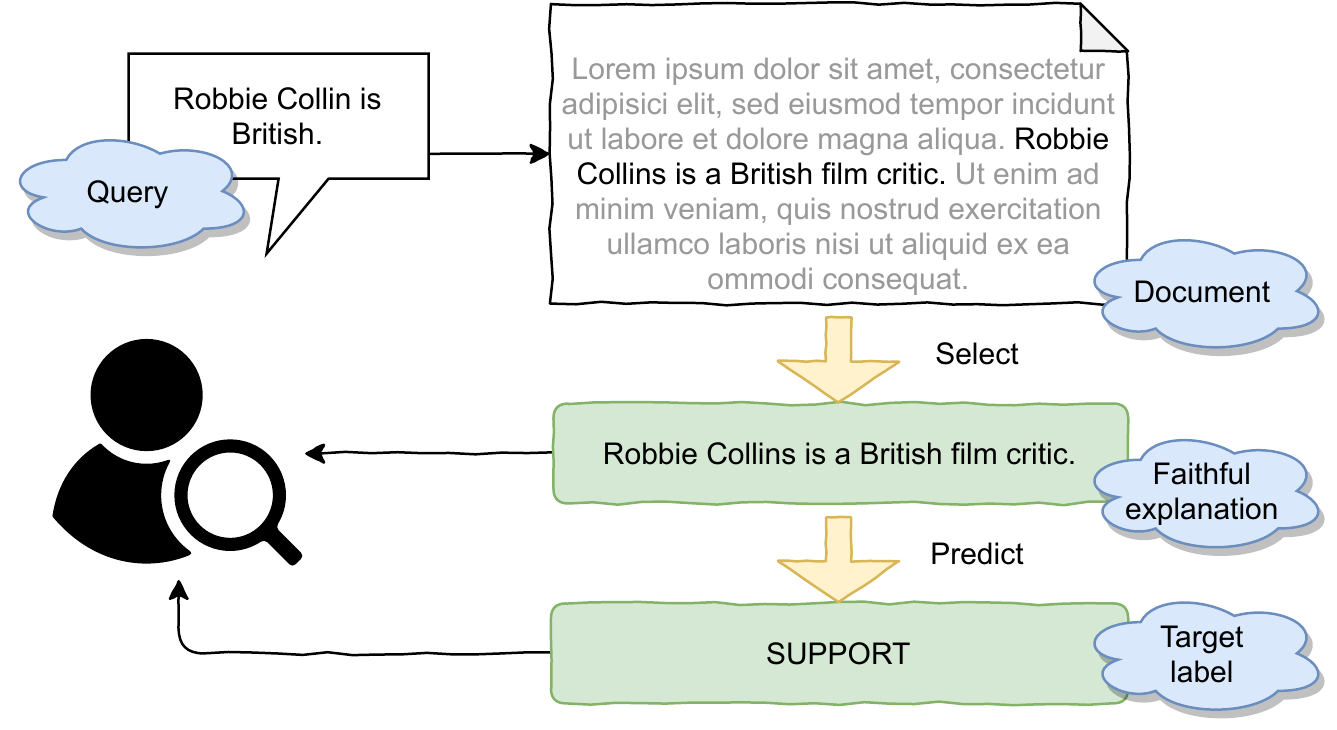}
      \caption{Example of the proposed rationale selecting process on one of the datasets (FEVER): Given a query and a document, our model selects the best rationale and predicts the label solely based on this selection.}
      \label{fig:main-components}
\end{figure}
Explainability has thus emerged as an increasingly popular field \citep{8631448,10.1145/3236009}.

We aim at faithful explanations -- the identification of the actual reason for \textit{the model's} prediction, which is essential for accountability, fairness, and credibility \citep{chakraborty2017interpretability,wu2018faithful} to evaluate whether a model's prediction is based on the correct evidence.
The recently published ERASER benchmark \citep{deyoung2019eraser} provides multiple datasets with annotated \emph{rationales}, i.e., parts of the input document, which are essential for correct predictions of the target variable  \citep{zaidan-etal-2007-using}.
By contrast to post-hoc techniques to identify relevant input parts such as LIME~\citep{ribeiro-2016-lime} or input~reduction~\citep{feng-etal-2018-pathologies}, we focus on models that are faithful by design, in which the selected rationale matches the full underlying evidence used for the prediction. 

Existing strategies mostly rely on \mbox{REINFORCE}~\citep{williams1992simple} style learning \citep{lei-etal-2016-rationalizing, yu-etal-2019-rethinking} or on training two disjoint models \citep{lehman-etal-2019-inferring, deyoung2019eraser}, in the latter case depending on rationale supervision. This poses critical limitations as rationale annotations are costly to obtain and, in many cases, not available. Additionally, only when the model can select the ``best" rationale from the full context we obtain an unbiased indicator for artifacts within a dataset that may influence models without rationale supervision.

In our proposed setup, we turn the hard selection into a differentiable problem by (a) decomposing each document into its residual sentences, and (b) similar to \citet{clark-gardner-2018-simple} optimize the weighted loss based of each of these candidates.
We show that this end--to--end trainable model (see Figure \ref{fig:main-components}) can compete with a standard BERT on two reasoning tasks without rationale--supervision, and even slightly improve upon it, when supervised towards gold rationales. Our quantitative analysis shows how we can exploit these extracted rationales to identify the model's decision boundaries and annotation artifacts of a multi--hop reasoning dataset.

\section{Related Work}
Understanding the deep neural networks' decisions has gained increasing interest in the research community \citep{deyoung2019eraser, alishahi2019analyzing, Wallace2019AllenNLP, jacovi2020towards}.
Several works are concerned with post--hoc techniques to explain decisions of blackbox models \citep{ribeiro-2016-lime, feng-etal-2018-pathologies,camburu2019can}. 
Visualizing attention weights has been heavily used, but is known to be insufficient \citep{jain-wallace-2019-attention, serrano-smith-2019-attention}.
Other works focus on making the models themselves more interpretable via neural module networks \citep{jiang-bansal-2019-self,nmn:iclr20}, graph--based networks \citep{tu2019select-answer-explain, xiao2019dynamically}, pipeline models \citep{lehman-etal-2019-inferring}, or by generating textual explanations \citep{NIPS2018_8163, rajani2019explain,liu-etal-2019-towards-explainable}. Rather than only producing this explanation as additional output, \citet{latcinnik2020explaining} base the target prediction on this automatically created hypothesis.

Some approaches jointly use rationales to explain the predictions and boost performance without ensuring faithfulness \citep{zaidan-etal-2007-using,melamud-etal-2019-combining, strout-etal-2019-human}. Recent work use Gumbel~Softmax~\citep{maddison2016concrete} to identify token--level rationales to avoid using \mbox{REINFORCE} \citep{bastings-etal-2019-interpretable, pfeiffer2019deep}.

Very recent work \citep{jain2020learning} aims similarly to us, to infer faithful rationales based on its impact on the target prediction without supervision, thereby relying on a dedicated explanation technique to identify rationales and an additional model for the prediction. This work is different in that we (a) rely on the same network weights for rationale selection \textit{and} target prediction, and (b) provide quantitative analysis about the decision criteria of the models on the reasoning tasks.

\section{Experimental Setup}
\subsection{Datasets}
\begin{figure}
\small
    \begin{tabularx}{\linewidth}{X}
    \toprule
    \textbf{FEVER} \\
    \textbf{Claim}\\
    \textit{Joan Crawford has had four marriages.} (SUPPORTS)\\
    \textbf{Document}\\
    {[}...{]} Following a public appearance in 1974 , after which unflattering photographs were published , Crawford withdrew from public life and became increasingly reclusive until her death in 1977 . \uline{\textsf{(R1)} Crawford married four times{~}.} \uline{\textsf{(R2)} Her first three marriages ended in divorce ; the last ended with the death of husband Alfred Steele .} Crawford 's relationships with her two older children , Christina and Christopher , were acrimonious . {[}...{]}\\
    \midrule
    \textbf{MultiRC} \\
    \textbf{Question}\\
    \textit{What are we seeing when we see lightning ?} \\
    \textbf{Answer}\\
    The discharge of electrons (TRUE) \\
    \textbf{Document}\\
    {[}...{]} Over time the differences increase . \uline{\textsf{(R1)} Eventually the electrons are discharged . This is what we see as lightning .} You can watch an awesome slow - motion lightning strike below . {[}...{]}\\
    \bottomrule
    \end{tabularx}
    \caption{While the example from FEVER provides two alternative single-sentence rationales (\textsf{R1} and \textsf{R2}), the MultiRC example requires considering two sentences at once for a single rationale (\textsf{R1}).}
    \label{tab:example-data}
\end{figure}{}
We conduct our experiments on three different datasets as provided by ERASER. Specifically, we use FEVER~\citep{thorne-etal-2018-fever}, MultiRC~\citep{khashabi-etal-2018-looking}, and Movies~\citep{zaidan-etal-2007-using} as shown in Table \ref{tab:datasets}.
We limit ourselves to this sub-set of ERASER, as they require the identification of rationales from multi--sentence documents (as opposed to single sentences). Further, our approach must process the full sample, including the document, within the same minibatch. We do not consider datasets if their documents' size imposes memory issues with pre--trained language models, as this would require external preprocessing, which is not controlled by the model.
\begin{table}[ht]
\small
    \centering
    \begin{tabular}{l c c c}
        & \textbf{FEVER} & \textbf{MultiRC} & \textbf{Movies}  \\
        \toprule
       \# Samples  &97,957 & 24,029 & 1,600\\
       Rationales / Sample & 1.0 & 1.5 &8.7\\
       \midrule
       \multicolumn{4}{l}{\textbf{Minimum reasoning--hops}}\\
       One  & 96,702 & - & 1,597\\
       Two  & 1,133 & 17,345 & -\\
       Three  & 73& 5,134 & 2\\
       Four  & 27&1,547 & -\\
       Five+  & 22& 3 & -\\
       \bottomrule
    \end{tabular}
    \caption{Properties of the datasets (train). In MultiRC rationales are annotated for each \textit{question}. The numbers here reflect counts per (\textit{question}, \textit{answer}) tuple.}
    \label{tab:datasets}
\end{table}{}
\paragraph{\textbf{FEVER}}
is a large fact--checking dataset based on Wikipedia. Given a claim and a relevant document, the model must either \textit{support} or \textit{refute} the claim\footnote{Note that this task--setup and dataset from \citet{deyoung2019eraser} differs from the original FEVER~\citep{thorne-etal-2018-fever}.}.
In FEVER, multiple alternative rationales may exist, each of which can be used to refute or support a claim. 
\paragraph{\textbf{MultiRC}} is a multi--hop--reasoning multiple--choice dataset. It encompasses a variety of genres. Each question is annotated with a single rationale, which always consists of multiple sentences. For each question, an arbitrary number of correct answers exists. Examples for both datasets can be found in Figure~\ref{tab:example-data}.

\paragraph{\textbf{Movies}} is a sentiment dataset of movie reviews. As opposed to the other two corpora,  
it (a) does not require reasoning between the document and an additional claim/question, and (b) contains rationale--annotations on a span--level. Though we are primarily interested in sentence--level reasoning tasks, we apply our method to this dataset and map its annotations to sentences.

\subsection{Our Model}
\paragraph{\textbf{Task Overview}}
We propose a model that (a) explains its decisions by outputting which input parts are used for the predictions as faithful rationales and (b) achieves performance comparable to a standard blackbox approach. Importantly, the model must be able to select rationales that are useful to solve the target task, without relying on additional supervision. We achieve this by first creating multiple smaller samples for each original sample --- each associated with a potential rationale --- and solving the task based on each sub--sample individually. Similar to \citet{clark-gardner-2018-simple}, each sub--sample is associated with a learned score. Our model utilizes this score to jointly predict the target and the rationale. Instead of learning these scores via direct supervision \citep{min-etal-2019-compositional}, our approach can derive them solely based on how useful each rationale is for solving the target task.
\begin{figure}
    \centering
    \includegraphics[width=\linewidth]{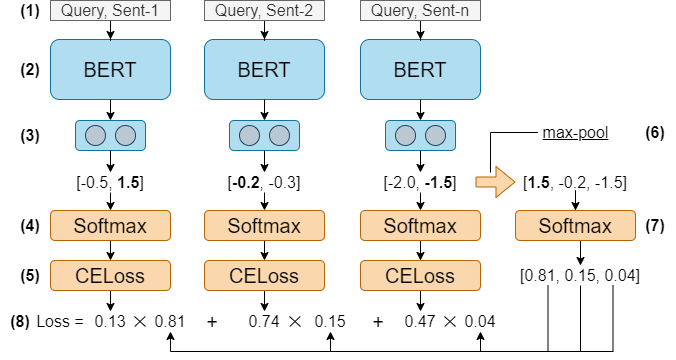}
    \caption{
    Model architecture. Each sample is split into its sentences (1), each individually encoded via BERT (2) followed by a linear layer (3). The loss for each input part is calculated separately (4,5). The score is computed via max--pooling (6), normalized (7) to compute the weighted loss (8). The input part with the highest score (6) is used for prediction.
    }
    \label{fig:approach-overview}
\end{figure}
\paragraph{\textbf{Single--Sentence without Rationale Supervision}}
\label{sec:single-sent-no-sup}
Given a sample, the model must predict the label $y$ based on a query $q$, i.e., the concatenation of the question and answer (MultiRC) or the claim (FEVER), and a document $D$. Instead of optimizing the objective given ($q$, $D$), we split $D$ into segments
and solve the overall task for each segment individually. We opt to split each document into sentences, as a trade-off between capturing enough semantic information within each segment while restricting each candidate's amount of information. Because some samples may be solved without any context \citep{schuster-etal-2019-towards}, we add a \textit{query--only} part, which is associated with no sentence ($\emptyset$). Hence, for each 
($q_k$, $D_k$) 
with $D_k$ containing $n_k$ sentences $s_{k,i}$, we create 
new input samples $x_k^{new}$ with $|x_k^{new}|=n_k+1$ as
\begin{equation}
\small
x_k^{new} = \bigg[(q_k, \emptyset), (q_k, s_{k,1}), (q_k, s_{k,2}), ..., (q_k, s_{k,n})\bigg]
\end{equation}
We use a standard model $\mathbf{m}$ to compute the logits $z_{k}$ (without softmax) based on all ($q_k$, $s_{k,i}$) in $x_k^{new}$ within the same minibatch.
All experiments use BERT-base-uncased~\cite{devlin2018bert} with a linear layer on top of the \texttt{[CLS]} token
\begin{equation}
\small
z_k=\mathbf{m}(x^{new}_k);\;\;\; z_k \in \mathbb{R}^{|x_k^{new}|\times t}
\end{equation}
whereas $t$ reflects the number of target labels.
Based on $z_k$ we compute $|x^{new}_k|$ losses $l_k$ via softmax and cross--entropy based on each 
($q_k$, $s_{k,i}$) 
individually. Likewise, $|x^{new}_k|$ different target predictions $\hat{y}_k$ are computed. Not all ($q_k$, $s_{k,i}$) contain the right information to properly solve the target task. Similar to \citet{clark-gardner-2018-simple, min-etal-2019-compositional} we rely on confidence scores to identify the best prediction, based on the most relevant rationale. To do so we must (a) compute scalar values $c_{k,i}$ as confidence scores for each ($q_k$, $s_{k,i}$), and (b) ensure that high scores $c_{k,i}$ are assigned to those input parts, that are most useful from \textit{the model's} perspective. We compute $c_k$ via row--wise max--pooling over $z_k$ as it represents the value of the selected class:
\begin{equation}
\small
c_k=\mathrm{max}(z_k)  \mathrm{;}\;\;\;c_k\in\mathbb{R}^{|x^{new}_k|} 
\end{equation}
The key idea is to multiply these $c_k$ with the losses $l_k$ to compute the overall loss, s.t. high losses will be associated with low confidence and vice-versa. Yet, we cannot merely multiply both terms, as this would allow the model to decrease the loss towards minus infinity only by assigning high negative values to all $c_k$ without optimizing towards the actual label.
To overcome this problem and obtain meaningful scores $c_k$ solely based on how useful each rationale is for the target task, we normalize all $c_k$ via softmax to obtain weights $w_{k,i}$ for each ($q_k$, $s_{k,i}$). As an overall objective, we minimize the weighted sum of losses using these weights:
\begin{equation}
\small
w_{k,i}=\frac{e^{\frac{c_{k,i}}{\tau}}}{\sum^{|c_k|}_{j=1}e^\frac{c_{k,j}}{\tau}}; \;\;\;
\mathrm{argmin}_\theta\bigg(\sum^{|x|}_{k=1}\sum^{|w_k|}_{i=1} w_{k,i}l_{k,i}\bigg)
\end{equation}
The rationale behind this is threefold: A right prediction, i.e., a low loss $l_{k,i}$, is only possible for informative sentences \textit{from the model's perspective}. First, by allowing the model to distribute the weights for the losses amongst all candidates, it can neglect non--informative sentences when learning to assign \textit{low} values (to high losses). Second, by normalizing these scores, it cannot ignore all sentences, but must assign comparatively \textit{higher} scores to at least one ($q_k$, $s_{k,i}$). Hence, to minimize the overall loss, high values must be assigned to the best suited ($q_k$, $s_{k,i}$), i.e., with the lowest (expected) loss. Finally, by deriving these scores directly from the predicted class, the same function for prediction and selection is used and optimized. The hyperparameter $\tau$ is the temperature of softmax, controlling the distribution of the softmax function. Higher values for $\tau$ result in a softer distribution, i.e., the loss is more evenly distributed amongst rationale candidates. Lower values result in a more hardened distribution, i.e., the model focuses quicker on one selected rationale.
For both, prediction and training, all rationales are always considered.
The process is visually exemplified in Figure~\ref{fig:approach-overview} and, for the most part (steps 2--5), resembles a standard setup.
\paragraph{\textbf{Prediction}}
For predictions, we select the sentence with the highest confidence from all sentences as the rationale $\hat{r}$, and the prediction based on $\hat{r}$ as the target $\hat{y}$:
\begin{equation}
\small
\hat{r} = \mathrm{argmax}(w) \mathrm{;}\;\; \hat{y}=\mathrm{argmax}(z_{\hat{r}})
\end{equation}
Though the rationale is \textit{faithful} on a sentence--level, we note that it does not indicate whether \textit{all} information of $\hat{r}$ is relevant to the model.

\paragraph{\textbf{Rationale supervision}}
We believe that rationales without supervision provide more trustworthy explanations. They are not affected by an additional objective and solely are selected if they are useful for the target task.
Nevertheless, we experimentally show how rationale--supervision can be applied by jointly \citep{yin-roth-2018-twowingos} supervising on the target and rationale.
To compute the rationale--loss as an additional objective, we treat slightly adapted confidence values $c_{k}^*$ as a multi--label problem via a sigmoid layer and binary cross--entropy loss.
\begin{equation}
\small
    c_{k,i}^*=
    \begin{cases}{}
    \mathrm{max}(z_{k,i}) & \text{if $x^{new}_{k,i}$ is not a gold--rationale.} \\
    z_{k,i,y} & \text{if $x^{new}_{k,i}$ is a gold--rationale.}\\
    \end{cases}
\end{equation}

\noindent This ensures that the correct class's confidence is increased even if the model (currently) predicts the wrong class.

\begin{table*}[ht]
    \small
    \centering
    \begin{tabular}{l | c c | c c c | c c}
         & \multicolumn{2}{c}{Target} & \multicolumn{3}{c}{Rationale} & \multicolumn{2}{c}{Target \& Rationale} \\
         & \textbf{F1a} & \textbf{Acc.} & \textbf{P} & \textbf{R} & \textbf{F1} & \textbf{Acc. Full}& \textbf{Acc. Part}\\
         \toprule
         \multicolumn{4}{l}{\textbf{FEVER} }\\
         Majority &33.2 & 49.6 &- & -& -& -& - \\
         BERT Blackbox & 90.2 \tiny{$\pm$0.4} & 90.2 \tiny{$\pm$0.4} & - & -& -  &  -  & -  \\
         Pipeline $\mathbb{S}$ \citep{deyoung2019eraser} & 87.7 & 87.8 & 88.3 & 87.7 & 88.0 & 78.1 & 79.0\\
         Single-Sentence Selecting $\mathbb{U}$ & 90.1 \tiny{$\pm$0.8} & 90.1 \tiny{$\pm$0.8}  & 80.0 \tiny{$\pm$4.3} & 79.4 \tiny{$\pm$4.3} & 79.7 \tiny{$\pm$4.3}& 72.2 \tiny{$\pm$4.5} &  73.2 \tiny{$\pm$4.5}\\
         Single-Sentence Selecting $\mathbb{S}$ & 90.7 \tiny{$\pm$0.7} & 90.7 \tiny{$\pm$0.7} & \textbf{92.3} \tiny{$\pm$0.1} & \textbf{91.6} \tiny{$\pm$0.1}& \textbf{91.9} \tiny{$\pm$0.1} &  \textbf{83.9} \tiny{$\pm$0.4} & \textbf{84.9} \tiny{$\pm$0.4} \\
         Two-Sentence Selecting $\mathbb{U}$ & 90.6 \tiny{$\pm$0.2} & 90.6 \tiny{$\pm$0.2} & 84.0 \tiny{$\pm$0.9} & 83.5 \tiny{$\pm$1.0}& 83.8 \tiny{$\pm$0.9} &  76.5 \tiny{$\pm$1.0} & 77.7 \tiny{$\pm$1.0} \\
         Two-Sentence Selecting $\mathbb{S}$ & \textbf{91.1} \tiny{$\pm$0.5} & \textbf{91.1} \tiny{$\pm$0.5} & 91.7 \tiny{$\pm$0.5} & 91.1 \tiny{$\pm$0.5}& 91.4 \tiny{$\pm$0.5} &  \textbf{83.9} \tiny{$\pm$0.8} & 84.8 \tiny{$\pm$0.7} \\
         \midrule
         \multicolumn{4}{l}{\textbf{MultiRC}}\\
         Majority &36.3 & 57.2 & - & - & -& -& - \\
         BERT Blackbox & 67.3 \tiny{$\pm$1.3} & 67.7 \tiny{$\pm$1.6} & - & - & - &  - & - \\
         Pipeline $\mathbb{S}$ \citep{deyoung2019eraser}& 63.3 & 65.0 & 66.7 & 30.2& 41.6& 0.0& 44.8\\
         Single-Sentence Selecting $\mathbb{U}$ & 65.2 \tiny{$\pm$3.5} & 66.8 \tiny{$\pm$3.8} & 34.6 \tiny{$\pm$24.5} & 15.5 \tiny{$\pm$10.9}& 21.4 \tiny{$\pm$15.1} &  0.0 \tiny{$\pm$0.0} & 23.3 \tiny{$\pm$16.6} \\
         Single-Sentence Selecting $\mathbb{S}$ & \textbf{67.4} \tiny{$\pm$0.4} & \textbf{69.1} \tiny{$\pm$1.3} & \textbf{74.3} \tiny{$\pm$1.1} & 33.5 \tiny{$\pm$0.5}& 46.1 \tiny{$\pm$0.6} &  0.0 \tiny{$\pm$0.0} & 54.0 \tiny{$\pm$0.9} \\
         Two-Sentence Selecting $\mathbb{U}$ & 66.7 \tiny{$\pm$2.7} & 67.7 \tiny{$\pm$3.0} & 44.4 \tiny{$\pm$11.0} & 19.9 \tiny{$\pm$5.0}& 27.5 \tiny{$\pm$6.9} &  0.1 \tiny{$\pm$0.0} & 31.2 \tiny{$\pm$7.4} \\
         Two-Sentence Selecting $\mathbb{S}$ & 65.5 \tiny{$\pm$3.6} & 67.7 \tiny{$\pm$1.5} & 65.8 \tiny{$\pm$0.2} & \textbf{42.3} \tiny{$\pm$3.9}& \textbf{51.4} \tiny{$\pm$2.8} &  \textbf{7.1} \tiny{$\pm$2.6} & \textbf{55.7} \tiny{$\pm$1.2} \\
         \midrule
         \multicolumn{4}{l}{\textbf{Movies}}\\
         Majority &33.3& 50.0& - & - & -& -& - \\
         BERT Blackbox &\textbf{90.1} \tiny{$\pm$0.3}& \textbf{90.1} \tiny{$\pm$0.3}& - & - & -& -& - \\
         Pipeline $\mathbb{S}$ \citep{deyoung2019eraser} &86.0 & 86.0 & \textbf{87.9} & 60.5 & 71.7& 40.7& \textbf{82.4} \\
         Single-Sentence $\mathbb{U}$ & 53.3 \tiny{$\pm$14.1} & 60.6 \tiny{$\pm$7.4} & 50.1 \tiny{$\pm$13.1} & 34.0 \tiny{$\pm$8.5}& 40.4 \tiny{$\pm$10.1} &  18.4 \tiny{$\pm$7.3} & 37.4 \tiny{$\pm$13.8} \\
         Single-Sentence $\mathbb{S}$ & 85.6 \tiny{$\pm$3.6} & 85.8 \tiny{$\pm$3.5} & 86.9 \tiny{$\pm$2.5} & \textbf{62.4} \tiny{$\pm$0.1}& \textbf{72.6} \tiny{$\pm$0.9} &  \textbf{43.9} \tiny{$\pm$0.6} & 81.4 \tiny{$\pm$3.9} \\
    \end{tabular}
    \caption{Mean performance and standard deviation for all models. $\mathbb{U}$ represents models without supervision on the rationale, $\mathbb{S}$ indicates supervision is applied on the rationale. The first two columns measure the performance on the target task using macro--averaged F1 and accuracy. The next three columns specify \underline{\textbf{P}}recision, \underline{\textbf{R}}ecall and \underline{\textbf{F1}} of the rationales on a sentence--level. The last two columns jointly show the performance based on a correct rationale \textit{and} target. Majority is only computed for the target--task performance.}
    \label{tab:results}
\end{table*}{}

\paragraph{\textbf{Multiple Sentences}}
Due to the memory consumption, encoding all (ordered) permutations of sentences up to a certain length through BERT is infeasible. To allow the model to select multiple sentences, 
for each permutation up to a length $h$, their representation is computed by max-pooling over the \texttt{[CLS]} token embeddings of its sentences.
We experiment with up to two sentences.

\section{Results}
\label{sec:results}
All experiments use AllenNLP~\citep{Gardner2017AllenNLP} and BERT-base-uncased~\citep{devlin2018bert} as provided by \citet{Wolf2019HuggingFacesTS}. We manually tune hyper-parameters for standard BERT baseline models and the sentence--selecting models, and show results in Table~\ref{tab:results}. We report results for the best configurations using three different seeds. 
We additionally report results of the BERT--to--BERT pipeline models from ERASER, which are based on the implementation of \citet{lehman-etal-2019-inferring}.
\paragraph{\textbf{Metrics}}
As opposed to \citet{deyoung2019eraser} we choose sentences as the lexical unit for rationales. We report precision, recall, and F1 for the rationales rather than token--level IOU, to avoid that the length of sentences impacts the metrics\footnote{To simplify comparisons with future work, we report the original ERASER metrics in Appendix~\ref{sec:appendix-eraser-metrics}.}.
As we are interested to understand whether a model makes the right prediction for the right reasons, we focus on \textit{sufficiency} of selected rationales rather than comprehensiveness: The claim of FEVER in Figure~\ref{tab:example-data} shows two valid rationales. Only one of these is required to support the claim. 
To compute precision, recall, and F1 w.r.t. sufficiency, we, therefore, compute these metrics based on the single, most similar\footnote{Determined by highest F1 on the sentence--level.} gold--rationale when evaluating any of the models.
We additionally report the joint accuracy of the target task and the rationale. Here we consider a prediction correct for the right reason, when it correctly predicts the target and \textit{all} sentences of one gold--rationale (Acc. Full). A weaker measure (Acc. Part) only requires the intersection of the selected sentences and one gold--rationale to be non--empty.
As multi--hop classification tasks tend to be easy to ``trick'' \citep{chen-durrett-2019-understanding}, this joint evaluation with the underlying evidence gives a better impression of the performance on the task itself.
\paragraph{\textbf{Observations}}
The \emph{Target} columns in Table~\ref{tab:results} show that
our models can compete with the standard BERT
on both reasoning tasks FEVER and MultiRC. This is especially surprising for single--sentence models on the multi--hop reasoning task MultiRC. We find that the single--sentence model $\mathbb{U}$ is more sensitive towards seeds, yielding in a slightly lower overall performance and higher variance on MultiRC (see Appendix~\ref{sec:appendix-seed sensitivity}). We believe this is because, given an unfortunate initialization, the model can focus on arbitrary features to quickly on this challenging dataset. Applying rationale supervision helps to stabilize this by improving the selected rationales rather than generally reaching higher target performances.  The BERT--to--BERT pipeline makes its prediction based on the best \textit{single} sentence and can only fairly be compared with the single--sentence selecting models.
The unsupervised approach is far behind all other models on the movies dataset, which we partly attribute to the small training data combined with the much larger document size. Primarily, however, we find (see Section~§\ref{sec:movies-analysis}) that by design, our approach is unsuitable for this kind of data, which due to its discussing nature, contains evidence for both labels within the same document.

The closest measure for ``right for the right reasons" is represented by Acc.~Full. 
Yet, it can only measure whether the prediction is based on the correct rationale on a sentence level, whereas it may still solely rely on certain contained words.
Assuming comprehensive rationale annotations\footnote{FEVER does not provide comprehensive rationale--annotations.}, the opposite can be said, i.e., 92.9\% of MultiRC are \underline{not} classified correctly for the right reasons. Note that both, the single--sentence models and the BERT--to--BERT pipeline, are bound to reach an 0\% for Acc.~Full on MultiRC, since they can only select a single sentence as the rationale.

\section{Analysis}
Leveraging the information about the used rationales, we closer analyze decision criteria for FEVER and MultiRC, and why our method performed poorly on Movies. Further, except for the two--sentence $\mathbb{S}$ models on MultiRC, no other model selects two sentences as a rationale in more than 1.3\%.
We partly attribute this to the less--than--optimal aggregation via max--pooling. 
As these are only selected due to the additional supervision, not for the utility to solve the overall task\footnote{We show supporting analysis for this in Appendix~\ref{sec:appendix-emultirc-multi-s-u}.},
we focus on single--sentence models.
\subsection{Poor Performance on Movies Dataset}
\label{sec:movies-analysis}
Without rationale supervision, our approach by far lacks behind its counterparts. To better understand the reason for this performance gap, we analyze the underlying data and the predictions. We find that our models $\mathbb{U}$ reach an average recall of $0.93$ and $0.32$ for \texttt{NEG} and \texttt{POS} respectively on the dev set --- despite the balanced training data. We emphasize that this is due to a very different nature of the data, compared to FEVER and MultiRC: Rather than all sentences within a document containing the same sentiment, they usually discuss pro and cons, and hence contain evidence for the gold label, as well as the opposite label. An extract of such a document can be seen in Figure~\ref{tab:example-movies} and two full examples in Appendix~\ref{sec:appendix-movies}. 
\begin{figure}
\small
    \begin{tabularx}{\linewidth}{X}
    \toprule
    \textbf{(35)} \underline{the scenes between nick and danny are very good}, and i actually got a feel for their characters; a bond forms between them that holds parts of the film together. \\
    \textbf{(36)} chow and wahlberg are both \underline{good actors}; chow is a pro, and can do this kind of stuff in his sleep. \\
    \textbf{(37)} wahlberg seems less at home in this atmosphere, but he's still \underline{fun to watch}. \\
    \textbf{(38)} \underline{i also liked the subplot} involving danny 's father; \underline{brian cox's performance is powerful}, and his character makes a compelling moral compass for danny. \\
    \textbf{(39)} but \underline{the film ultimately fails}, mostly at the hands of insane incoherence and overly - familiar action scenes. \\
    \bottomrule
    \end{tabularx}
    \caption{An extract of a movie review with an overall \textbf{negative sentiment}. Sentence 35--38 in isolation contain positive sentiment, whereas sentence 39 shows strong negative sentiment. Only the underlined span in line 39 constitutes a gold rationale and represents the overall sentiment.}
    \label{tab:example-movies}
\end{figure}{}
During prediction, even for humans, it is impossible to predict the correct overall sentiment based on isolated, out of context sentences of opposing stances. An additional problem arises during training in our setup: For the presented example, the model must either learn to either predict the label \texttt{NEG} even for sentences with clearly (only) positive indicators, or learn to reduce their confidence values $c_k$ to mitigate their impact. Either way, this naturally compromises its ability to detect the opposite sentiment. This discussion--based nature of Movies significantly differs from MultiRC and Fever. In the latter case, each document only contains evidence for \textit{or} against a claim, not both. In this case, the model must not learn contradicting patterns and only lower the confidence for irrelevant sentences, consistent with both labels. Both, the pipeline and the model $\mathbb{S}$, show that by guiding the model towards gold rationales, it can detect sentences for the overall movie sentiment. Without this guiding, however, our approach seems not suitable for such tasks.

\subsection{Learning curves}
\begin{figure}
    \centering
    \includegraphics[width=\linewidth]{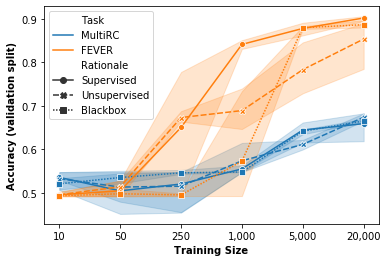}
    \caption{Accuracy (validation split) of BERT Blackbox and single--sentence models by training sizes.}
    \label{fig:acc_by_data}
\end{figure}

We investigate the impact of the amount of available training data for the three different models blackbox, model $\mathbb{S}$, and $\mathbb{U}$.
To limit the data's impact, we create three random subsets of the training data of different sizes and report the average performance of each of the models on these subsets in Figure~\ref{fig:acc_by_data}. 
{}
All three models show similar trends across all training sizes for MultiRC. On FEVER, the rationale--supervision offers an additional boost in scenarios with little data. 
Without rationale--supervision, it tends to require more data to reach its peak performance.
\subsection{Model decisions on FEVER}
\begin{figure}
    \centering
    \includegraphics[width=\linewidth]{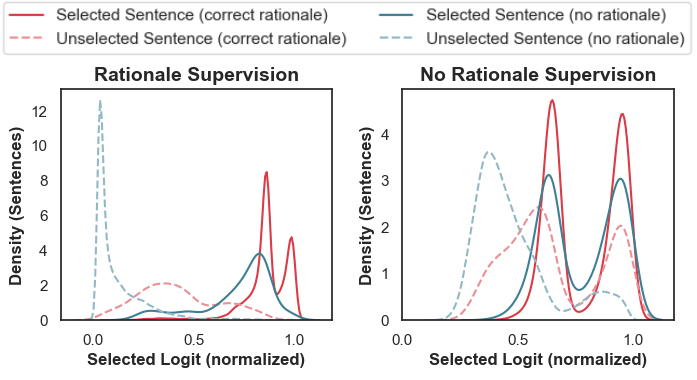}
    \caption{KDE plots of the single--sentence models ($\mathbb{S}$ left) and ($\mathbb{U}$ right) for FEVER, showing  the relative frequency for each category individually based on globally normalized logits $z_{k,i}$ of the selected label.}
    \label{fig:fever-logits-by-rationales}
\end{figure}{}
Both (best) single--sentence models $\mathbb{U}$ and $\mathbb{S}$ perform very strong and predict the same label in 93.8\% of all cases, from which they select the same rationale in 86\%. We, therefore, focus on how supervision affects the model internally. 
Specifically, we exploit the fact that relevance and prediction are jointly encoded and optimized within the same logits $z_{k,i}$.
In Figure~\ref{fig:fever-logits-by-rationales} we compare these $z_{k,i}$
from a global perspective after normalizing them using min--max--normalization.
Applying rationale--supervision leads to more decisive predictions, as the vast majority of unselected sentences scores close to the global minimum, whereas selected sentences have scores close to the maximum. Invalid selected rationales tend to be shifted slightly more towards the lower end than selected correct rationales. This looks very different for model~$\mathbb{U}$. Most importantly, a non--trivial amount of unselected sentences reached scores very close to the global maximum.

\paragraph{\textbf{Does it learn semantically better decision criteria with supervison?}}
A possible reason why such high values occur for unchosen sentences is that the selected rationale is not substantial for a correct target prediction. \citet{schuster-etal-2019-towards} identify n--grams within claims that highly correlate with certain classes.
By adding new evidence and claims for each of their selected claims they design a symmetric test-set, which cannot be solved using such artifacts. Intuitively, 
similar to \citet{stacey2020there}, 
applying rationale--supervision (model $\mathbb{S}$) forces the model to learn --- based on the rationale --- high and low values for the same claim, i.e. containing the same artifacts. It should therefore be more sensitive for the context and not rely on claim--only features. We show the performance on this symmetric test set in Table~\ref{tab:fever-sym}.
\begin{table}[]
    \small
    \centering
    \begin{tabular}{l c c c}
         &  \textbf{BERT} & \textbf{Single $\mathbb{U}$} & \textbf{Single $\mathbb{S}$}\\
         \toprule
        F1 (SUPPORT) & 67.8 \tiny{$\pm$0.6}& 68.1 \tiny{$\pm$0.4}& \textbf{71.1} \tiny{$\pm$1.9}\\
        F1 (REFUTE) & 61.3 \tiny{$\pm$2.4}& 62.1 \tiny{$\pm$0.7}& \textbf{64.4} \tiny{$\pm$2.4}\\
        F1a & 64.5 \tiny{$\pm$1.5}& 65.1 \tiny{$\pm$0.2}& \textbf{67.8} \tiny{$\pm$3.7}\\
        \bottomrule
    \end{tabular}
    \caption{Evaluation of BERT and single--sentence selecting models on the symmetric FEVER testset \citep{schuster-etal-2019-towards} (717 samples)}
    \label{tab:fever-sym}
\end{table}
Despite a small improvement, it still lacks far behind the performance on FEVER. Even the model $\mathbb{U}$ rarely selects the claim--only as the rationale, suggesting that at least partially, additional context helps to solve the task properly. Yet, it shows that smaller lexical units than sentences as a rationale may be beneficial in such cases.
\subsection{Model decisions on MultiRC}
\label{sec:multirc-analysis}
\subsubsection*{What is the impact of rationale supervision?}
The ceiling performance on the target task remains the same, even with rationale--supervision. We analyze the validity of the selected rationales on the validation split to shed light on (a) how the model can achieve a strong performance, and (b) how rationale supervision affects the model. For simplicity, we select the best performing single--sentence models and group the predictions by the gold and predicted target label in Table~\ref{tab:multirc-absence-of-evidence}.
\begin{table}[]
\small
    \centering
    \begin{tabular}{l|c c c c}
         &  \textbf{T-T} & \textbf{T-F} & \textbf{F-T} & \textbf{F-F} \\
         \toprule
    $\mathbb{U}$ Rationale Prec. & 79.4 & 62.3 & 45.9 & 36.2 \\
    $\mathbb{S}$ Rationale Prec. & 86.5 & 78.2 & 52.4 & 80.3 \\
    \midrule
        $\Delta$-Rationale Prec. &\textbf{+7.1} &\textbf{+15.9} &\textbf{+6.5} &\textbf{+44.1} \\
        \bottomrule
    \end{tabular}
    \caption{Precision of the selected rationale by the best single--sentence models on MultiRC, grouped by the (Gold - Predicted) labels \underline{\textbf{T}}rue and \underline{\textbf{F}}alse.}
    \label{tab:multirc-absence-of-evidence}
\end{table}{}
The model $\mathbb{U}$ results show that evidence of positive samples is more likely to get selected. While the correctly predicted positive samples mostly rely on gold \textit{evidence for} the answer, for correctly predicted negative samples, the \textit{absence of supporting evidence} seems sufficient, rather than explicit evidence against it. Note that none of these ``evidence" is truly sufficient, as multiple sentences are technically required. To see whether this behavior is due to our training method or helpful for the underlying data, we re--evaluate the best performing BERT on the validation set and exclude all gold--rationales from the documents. The results show a recall of $28.4$ (True) and $81.8$ (False)\footnote{Compared to 54.5 (True) and 79.3 (False)}, suggesting a similar behavior. Hence, the major benefit from rationale--supervision is to predict the label \texttt{False} based on the correct sentence, which is not required to solve the overall task. To limit this property of future datasets, we 
believe it is important to add unanswerable instances,
as done for instance by \citet{thorne-etal-2018-fever} or \citet{rajpurkar-etal-2018-know}.

\subsubsection*{What kind of sentences are selected as a rationale?}
\begin{figure}
    \centering
    \includegraphics[width=\linewidth]{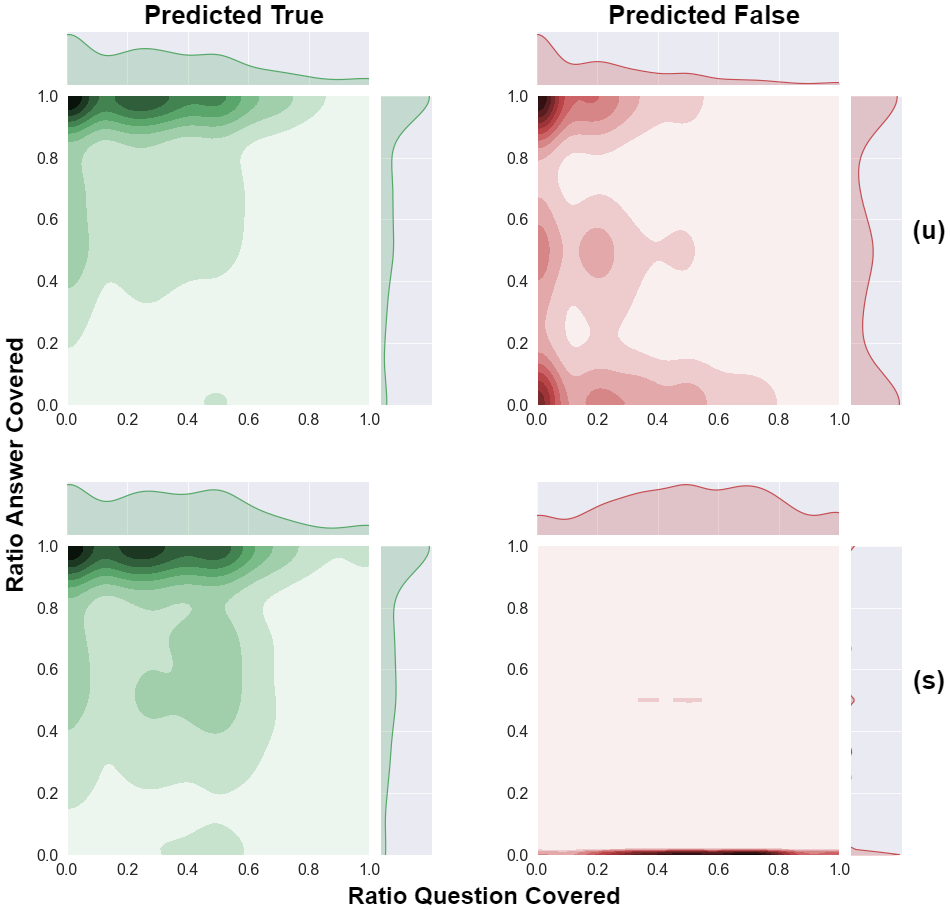}
    \caption{KDE plots for word overlaps between Question/Answer and the selected rationale of single--sentence models on MultiRC with (bottom) and without (top) rationale supervision..}
    \label{fig:overlap-multirc-u-s}
\end{figure}

We jointly look at the selected sentences with the target prediction of both models $\mathbb{U}$ and $\mathbb{S}$ and observe a high correlation with word overlap of the question and 
the answer. Figure~\ref{fig:overlap-multirc-u-s} shows KDE plots of the selected sentences based on the percentage of non--stopwords\footnote{We use spaCy to exclude punctuation and stopwords and seaborn~\citep{michael_waskom_2017_883859} with default parameters for plotting.} of the question and answer respectively, that are also contained within the selected sentence. 
{}
We make multiple observations: Positive predictions mostly depend on a high overlap with the answer. The overlap with the question has a lower priority. Especially for the model $\mathbb{S}$, a clear decision boundary between rationales for both labels can be seen based on the lexical overlap. Interestingly, also \citet{yadav-etal-2019-quick}, to a large part, rely on similar lexical features for their unsupervised detection of justification sentences on MultiRC. 
In line with the previous section, rationale supervision only has a limited impact on positive predictions.
A significant difference is shown for the negative predictions. 
Whereas model~$\mathbb{U}$ tends to select rationales for both labels based on similar criteria, the selected rationales for samples predicted \texttt{False} by model~$\mathbb{S}$ almost entirely have lexical overlaps with the question only.
This intuitively makes sense, as the same rationales are valid for each question. Negative rationales should therefore be relevant for the question, not for the answer. We show some examples in Appendix~\ref{sec:multirc-examples}.
\subsubsection*{Are single sentences sufficient for MultiRC?}
It has been shown that noisy detection of evidence can already improve the performance on MultiRC \citep{wang-etal-2019-evidence}, yet this should not be possible via single sentences.
To see whether BERT exploits such biases, we follow \citet{gururangan-etal-2018-annotation} and identify samples within the test--set that are solvable using a single--hop only, i.e., these which the single--sentence $\mathbb{U}$ model classified correctly. To limit the impact of lucky guesses, we group samples by the number of these models that could solve them in Table~\ref{tab:multirc-hard-easy}. 
As pointed out in Section~\ref{sec:results}, one of our single--sentence models $\mathbb{U}$ on MultiRC performed poorly due to its seed sensitivity . To exclude impacts from this specific model and group the test--split by meaningful criteria, we retrain BERT blackbox and model~$\mathbb{U}$ with a new random seed, reaching an F1a score of 66.3 and 67.6 respectively on the test set. We select the best three seeds of both model types for splitting the data (model~$\mathbb{U}$) and evaluation (BERT blackbox).
\paragraph{\textbf{Lexical Overlap Logistic Regression}}
Additionally, we mimic our observations with the high lexical overlap using a simple logistic regression. We calculate a rationale score $r=w_qq_s + w_aa_s$ for each sentence $s$,
whereas $q_s$ and $a_s$ represent the absolute/relative word overlap of the sentence with the question and answer respectively. For each sample, the sentence with the highest $r$ is selected as a rationale (shorter sentences are preferred as a tie--breaker) and used to train a logistic regression (LR), breaking down the multi--hop reasoning task to two digits based on a single sentence. We run a grid--search with different values for $w_q$ and $w_a$ and select the model with the highest F1a score of 63.5 on the validation set (F1a score of 58.1 on the test set), using absolute word overlaps, $w_q=0.4$ and $w_a=1.0$.
\paragraph{\textbf{Results}}
The performances are shown in Table~\ref{tab:multirc-hard-easy}. 
\begin{table}[]
\small
    \centering
    \begin{tabular}{l c c c c}
       & \textbf{3/3} & \textbf{2/3} & \textbf{1/3} & \textbf{0/3} \\
       \toprule
       \textbf{Size}  & 2,314 & 1,114 & 779 & 641 \\
       \midrule
       \multicolumn{5}{l}{\textbf{Logistic Regression (F1)}}\\
       \textbf{True}  & 70.2 & 61.4 & 59.6 & 48.6 \\
       \textbf{False}  & 69.2 & 29.0 & 15.5 & 9.1 \\
       \textbf{F1a}  & 69.7 & 45.2 & 37.5 & 28.8 \\
       \midrule
       \multicolumn{5}{l}{\textbf{BERT Blackbox (F1)}}\\
       \textbf{True}  & 91.7\tiny{$\pm$0.9} & 62.3\tiny{$\pm$3.5} & 42.3\tiny{$\pm$5.0} & 14.8\tiny{$\pm$3.5} \\
       \textbf{False}  & 94.9\tiny{$\pm$0.6} & 69.1\tiny{$\pm$1.5} & 45.5\tiny{$\pm$5.9} & 8.7\tiny{$\pm$4.3} \\
       \textbf{F1a}  & 93.3\tiny{$\pm$0.8} & 65.7\tiny{$\pm$1.0} & 43.9\tiny{$\pm$1.7} & 11.8\tiny{$\pm$1.4} \\
       \textbf{$\Delta$F1a} & \textbf{+25.3}\tiny{$\pm$1.1}& \textbf{--2.3}\tiny{$\pm$0.8} & \textbf{--24.1} \tiny{$\pm$1.6} & \textbf{--56.2} \tiny{$\pm$1.4}\\
       \bottomrule
    \end{tabular}
    \caption{Average performance of BERT models based on subsets of the test--split that can be solved using a single sentence, compared with a lexical overlap logistic regression. $\Delta$F1a measures the difference w.r.t. the performance on the full test set. Columns indicate how many single--sentence models $\mathbb{U}$ could solve each contained instance correctly.}
    \label{tab:multirc-hard-easy}
\end{table}
BERT performs strongly on samples that can be solved using a single sentence while struggling with the same instances as model~$\mathbb{U}$.
Further, a simple logistic regression shows a similar trend. On the easiest (and largest) part it even exceeds the performance of the full test--set of any BERT model. The results suggest that
high performance does not indicate successful multi--hop reasoning\footnote{This is not the official, hidden test--set of MultiRC.}.

\section{Discussion}
\paragraph{\textbf{Limitations}}
From a technical perspective, a limitation is memory consumption, as the model must process all rationale candidates of the same instance within the same minibatch. Though single--sentence rationale can be processed, encoding all combinations of multiple sentences via BERT is problematic.
Future work could investigate better sampling strategies or a greedy breadth search to reduce the number of candidates. 
Another limitation is the inability of coreference resolution between different sentences and the consideration of the context in general. 
Solving this is non--trivial, as we essentially buy faithfulness by explicitly omitting all other information than the selected sentence(s). While this does not seem crucial in the evaluated datasets, it poses potential dangers for malicious attacks, most importantly, when considering the permutations of multiple sentences. Therefore, we recommend to always show the identified evidence in context when using our approach in the real world.

\paragraph{\textbf{Conclusion}}
We proposed a conceptually simple approach to allow models to extract faithful rationales, which can compete with standard BERT on two reasoning tasks without supervision and even improve the overall performance, when supervising on the rationale. We showed that by outputting faithful rationales, it is possible to not only compare models based on the target performance alone, but also quantify how well even those correct predictions are based on the correct evidence.
Our analysis showed that exploiting this knowledge about the selected rationales helps shed light on the models' the decision mechanism for debugging purposes and on the underlying data.

\section*{Acknowledgments}
We thank the anonymous reviewers for their constructive feedback.
This research work has been funded by the German Federal Ministry of Education and Research and the Hessen State Ministry for Higher Education, Research and the Arts within their joint support of the National Research Center for Applied Cybersecurity ATHENE, and the``Data Analytics for the Humanities" grant by the Hessian Ministry of Higher Education, Research, Science and the Arts. 
We gratefully acknowledge the support of NVIDIA Corporation with the donation of the Titan X Pascal GPU and the Titan Xp Pascal GPU used for this research.

\bibliography{emnlp2020}
\bibliographystyle{acl_natbib}

\newpage
\appendix
\label{sec:appendix}

\section{ERASER Metrics}
\label{sec:appendix-eraser-metrics}
\begin{table}[h!]
\small
    \centering
    \begin{tabular}{l c c c }
         &  \textbf{F1a} & \textbf{IOU F1} & \textbf{Token F1}\\
         \toprule
         \textbf{FEVER} & & & \\ 
       \citeauthor{lei-etal-2016-rationalizing} $\mathbb{U}$  &71.8 &0.0 & 0.0\\ 
       \citeauthor{lei-etal-2016-rationalizing} $\mathbb{S}$  &71.9 & 21.8& 23.4\\ 
       \citeauthor{deyoung2019eraser} $\mathbb{S}$  & 87.7& 83.5& 81.2\\ 
       Single--Sentence $\mathbb{U}$ & 90.1 \tiny{$\pm$0.8}&75.6 \tiny{$\pm$4.0} &73.7 \tiny{$\pm$3.9} \\
       Single--Sentence $\mathbb{S}$ & 90.7 \tiny{$\pm$0.7}&\textbf{87.3} \tiny{$\pm$0.1} &\textbf{85.1} \tiny{$\pm$0.1} \\
       Two--Sentence $\mathbb{U}$ & 90.6 \tiny{$\pm$0.2}&79.5 \tiny{$\pm$0.9} &77.5 \tiny{$\pm$0.8} \\
       Two--Sentence $\mathbb{S}$ & \textbf{91.1} \tiny{$\pm$0.5}&86.8 \tiny{$\pm$0.5} &84.6 \tiny{$\pm$0.5} \\
       \midrule
       \textbf{MultiRC} & & & \\ 
       \citeauthor{lei-etal-2016-rationalizing} $\mathbb{U}$  &64.8 &0.0 &0.0 \\ 
       \citeauthor{lei-etal-2016-rationalizing} $\mathbb{S}$  &65.5 &27.1 & 45.6\\ 
       \citeauthor{deyoung2019eraser} $\mathbb{S}$  &63.3 &41.6 & 41.2\\ 
        Single--Sentence $\mathbb{U}$ & 65.2 \tiny{$\pm$3.5}&21.4 \tiny{$\pm$15.1} &20.9 \tiny{$\pm$14.8} \\
       Single--Sentence $\mathbb{S}$ & \textbf{67.4} \tiny{$\pm$0.4}&46.1 \tiny{$\pm$0.6} &45.0 \tiny{$\pm$0.7} \\
       Two--Sentence $\mathbb{U}$ & 66.7 \tiny{$\pm$2.7}&27.5 \tiny{$\pm$6.9} &27.7 \tiny{$\pm$6.7} \\
       Two--Sentence $\mathbb{S}$ & 65.5 \tiny{$\pm$3.6}& \textbf{51.4} \tiny{$\pm$2.8} & \textbf{49.0} \tiny{$\pm$2.6} \\
       \midrule
       \textbf{Movies} & & & \\ 
       \citeauthor{lei-etal-2016-rationalizing} $\mathbb{U}$  &\textbf{92.0} &1.2 &32.2 \\ 
       \citeauthor{lei-etal-2016-rationalizing} $\mathbb{S}$  &91.4 &\textbf{12.4} &\textbf{28.5} \\ 
       \citeauthor{deyoung2019eraser} $\mathbb{S}$  & 86.0& 7.5& 14.5\\ 
              Single--Sentence $\mathbb{U}$ & 53.3 \tiny{$\pm$14.1}&3.2 \tiny{$\pm$1.3} &7.8 \tiny{$\pm$2.5} \\
       Single--Sentence $\mathbb{S}$ & 85.6 \tiny{$\pm$3.6}&7.0 \tiny{$\pm$0.2} &15.3 \tiny{$\pm$0.3} \\
       \bottomrule
    \end{tabular}
    \caption{Results on the original ERASER metrics together with their reported performance using the REINFORCE approach by \citet{lei-etal-2016-rationalizing} and the BERT--to--BERT pipeline by \citet{deyoung2019eraser}.}
    \label{tab:eraser-original}
\end{table}
\section{MultiRC Sensitivity to Seeds}
\label{sec:appendix-seed sensitivity}
\begin{table}[h!]
\small
    \centering
    \begin{tabular}{l c c c c}
    \textbf{Model} & \textbf{F1a} & \textbf{Acc}& \textbf{Rat. P.}& \textbf{Acc. Part}\\
    \toprule
        Single--Sent--1 $\mathbb{U}$ &69.4 &69.9 & 53.1& 37.1\\
        Single--Sent--2 $\mathbb{U}$ & 60.9 &61.3 &0.0 & 0.0\\
        Single--Sent--3 $\mathbb{U}$ & 65.5& 69.0&50.7 &32.8 \\
        \midrule
        Blackbox--1 &67.9 &69.0 &- & -\\
        Blackbox--2 &68.6 &68.4 &- & -\\
        Blackbox--3 &65.4 &65.5 &- & -\\
        \bottomrule
    \end{tabular}
    \caption{Performance of BERT blackbox models and Single--Sentence $\mathbb{U}$ models across different seeds on MultiRC.}
    \label{tab:multirc-single-u-seeds}
\end{table}
\newpage
\section{Examples for MultiRC with and without supervision}
\label{sec:multirc-examples}
We show representative samples for both gold labels with the same and distinct target predictions. In cases where only one model is correct, \texttt{True} labelled samples are mostly classified correctly by $\mathbb{U}$ (82.7\%), \texttt{False}-labelled samples by $\mathbb{S}$ (81.5\%). We decide whether to show distinct or same rationales, depending on the majority of cases within each of these categories.
\begin{figure}[hb]
    \centering
    \includegraphics[width=\linewidth]{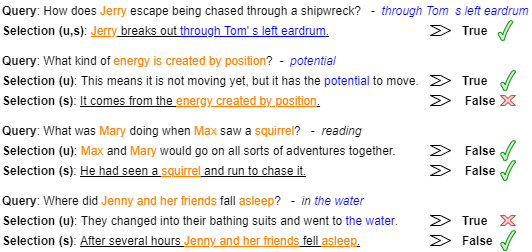}
    \caption{Examples from MultiRC with selected rationales and their prediction for the single-sentence model with $\mathbb{S}$ and without $\mathbb{U}$ rationale--supervision. \underline{Underlined} sentences are part of the gold--rationale, word--overlaps are highlighted with colors.}
    \label{fig:multirc_single_examples}
\end{figure}{}

\section{Two-Sentence Models on MultiRC with and without supervision}
\label{sec:appendix-emultirc-multi-s-u}
\begin{table}[h!]
\small
    \centering
    \begin{tabular}{l | c c}
    & \multicolumn{2}{c}{\textbf{Prediction both sents}} \\
        \textbf{\% Same Prediction}  &  \textbf{False} &  \textbf{True}\\
         \toprule
        \textbf{Sentence (Shared)} & 79.4\%& 96.8\%\\
         \textbf{Sentence (New)} & 99.5\% & 51.3\%\\
    \end{tabular}
    \caption{Change of target prediction based on single sentences of model~$\mathbb{S}$, when identifying two sentences as rationale. Columns indicate the classification based on the identified rationale. Rows show how many of these instances are still classified the same, when only using the same single sentence as rationale, as used by model~$\mathbb{U}$ (\textbf{Shared}), or by the additional sentence, only selected with rationale--supervision (\textbf{New}).}
    \label{tab:multirc-u-vs-s-table}
\end{table}
On MultiRC, the two--sentence model $\mathbb{U}$ selects a single sentence as the rationale in 99.0\%, whereas the model~$\mathbb{S}$ selects two sentences on 51.4\%. In 83.4\% both models predict the same target $\hat{y}$. Based on these, we consider all instances, where model~$\mathbb{S}$ selects the same sentence as model~$\mathbb{U}$ plus one additional sentence as a rationale, to identify whether (i) both sentences are relevant, (ii) the shared sentence is relevant, or (iii) the additional sentence is relevant for model~$\mathbb{S}$. Instead of looking at the prediction of the joint rationale of both sentences of model~$\mathbb{S}$, i.e., the selected rationale with the highest confidence score, we now look at the predictions of both selected sentences individually. Table~\ref{tab:multirc-u-vs-s-table} shows whether the prediction of model~$\mathbb{S}$ remains stable for both predicted labels if only one of the sentences out of the two--sentence rationale is used.
For \texttt{False} predictions, the additional sentence (only selected when supervised) has a major impact on the prediction and seems most relevant. This is in line with our observations in Section~\ref{sec:multirc-analysis}, namely that supervision affects the decision mechanism predicting this label. For the prediction of \texttt{True}, in almost all cases the same sentence as the one selected by model~$\mathbb{U}$ yields in the same prediction. The additional sentence in isolation, however, changes the prediction to \texttt{False} in almost half of all cases. Though bound to our approach, these results suggest that rationale--supervision may yield in selecting rationales that are not required by the model to solve the target task, but rather the rationale--objective
, thereby losing some of their faithfulness. This may be a relevant consideration when measuring faithfulness on a more fine--granular level.

\section{Movies Examples}
Figure~\ref{tab:example-appendix-1} shows an example of positive sentiment in which the model disregards sentences with clear positive stances and selects a sentence containing ``scary" as the rationale. Figure~\ref{tab:example-appendix-2} shows how the model correctly selects a sentence of positive stance but interprets this sentence as negative. Both examples show that sentences with opposing stances occur by discussing the plot and the movie in general.
\label{sec:appendix-movies}
\begin{figure*}
\small
    \begin{tabularx}{\linewidth}{X}
    \toprule
\textbf{(1)}  there ' s a thin line between satire and controversy , and mike nichols ( the birdcage , wolf ) has directed a sharp and very honest look at a us presidential election . \\
\textbf{(2)}  based on the book written by " anonymous " ( actually former " newsweek " writer joe klein ) , john travolta plays governor jack stanton . \\
\textbf{(3)}  but he does n ' t actually play stanton . \\
\textbf{(4)}  he plays bill clinton ; just the same as emma thompson no doubt plays the first lady and billy bob thorton is the campaign manipulator james carville ( although the credits will of course say otherwise ) . \\
\textbf{(5)}  the film is taken from the perspective of henry burton ( adrian lester ) , a morally correct and somewhat hesitant new advisor to stanton . \\
\textbf{(6)}  he searches for justice and dignity in the ugliest possible situations , and whether it be keeping the history of his boss ' pants under wraps or contemplating digging up dirt on another politician , he approaches his work with a keen desire to skillfully serve his country and his fellow workers . \\
\textbf{(7)}  richard jemmons ( billy bob thorton ) and daisy green ( maura tierney ) team up with henry as the would - be president ' s advisors , and hire lesbian veteran libby holden ( kathy bates ) as the campaign ' s eccentric " tougher than dirt " incriminator . \\
\textbf{(8)}  together they face all sorts of sexual allegations , the irritatingly discourteous media and other witty politicians in the election race . \\
\textbf{(9)}  in its satire and controversy , primary colors is a similar film to wag the dog : they both are not afraid to wipe their noses in the nitty - gritty and take a bold look at something that will never has honesty as a virtue . \\
\textbf{(10)}  but whereas wag showed us how much affect a few people can have on the media , primary colors is much more concerned with fleshing out it ' s characters , letting us understand what they want and why , and making us truly appreciate the humanity and rectitude that they graciously represent . \\
\textbf{(11)}  seeing john travolta play bill clinton \\
\textbf{(12)} \textit{ so confidently and justly is enough to make the film more than worth a look . and the rest of the cast also make }\\
\textbf{(13)} \textit{ superb performances - adrian lester sharply portrays the intellect of henry whilst kathy bates is perfect as the robust and energetic libby holden . }\\
\textbf{(14)}  at occasions , you ca n ' t help but feel that these terrific characters are going to waste . \\
\textbf{(15)}  there are long slabs of time where john travolta ( unquestionably the most interesting to watch ) is missed from the screen ; and since it is awkwardly structured as henry ' s story we are often forced to watch scenes that perhaps are not so necessary to the central plot - or even the point of the film . \\
\textbf{(16)} \textit{ having said that , make no mistake - primary colors is always enjoyable to watch }\\
\textbf{(17)}  . \\
\textbf{(18)}  but frequently we have to ask ourselves - exactly what are we watching ? \\
\textbf{(19)}  most of the first half of its duration is a lightheaded look at melodramatic confrontations that seem so genuine we can not help but laugh , but the way primary colors chooses to finish tackles aspects that are very contrary , and almost unsuitable , to the rest of the film . \\
\textbf{(20)} \textit{ but as i mentioned before , there is a thin line between satire and controversy - and for the most part , primary colors delivers an entertaining indulgence of political matters combined with a far - from - overpowering look at winning the public ' s opinion . }\\
\textbf{(21)}  although at occasions the film may jump around a little too freely , focus is never lost on how important and vulnerable the subject matter really is . \\
\textbf{(22)}  thankfully , it is clear to make the distinction on what is entertaining movie cosmetics and what is a provocative documentation of something \\
\textbf{(23)} \textbf{ so really it ' s scary . }\\
    \bottomrule
    \end{tabularx}
    \caption{Example of Movies (dev) with gold label \texttt{POS} and predicted label \texttt{NEG}. \textit{Italic} sentences are gold (sentence--level) rationales, \textbf{bold} is the selected rationale. }
    \label{tab:example-appendix-1}
\end{figure*}{}

\begin{figure*}
\small
    \begin{tabularx}{\linewidth}{X}
    \toprule
\textbf{(1)}  buffalo ? \\
\textbf{(2)}  66 is a very rarely known movie that stars vincent gallo and christina ricci . \\
\textbf{(3)}  gallo plays a very troubled man , who was sent to jail for gambling . \\
\textbf{(4)}  once out of jail , he must visit his parents , who he told he was married . \\
\textbf{(5)}  the truth is he is n ' t married . \\
\textbf{(6)}  to try to impress them , he kidnaps a girl ( christina ricci ) from a tap dancing class to act as his wife . \\
\textbf{(7)}  the film is very cheaply made , and it shows it throughout a lot of the movie , but you do n ' t need money to make a good film . \\
\textbf{(8)}  buffalo ? \\
\textbf{(9)}  66 does n ' t always stay with the realistic concept , and sometimes goes through outrageous events . \\
\textbf{(10)}  gallo ' s parents , played by angelica huston and ben gazarra , are two very strange individuals . \\
\textbf{(11)}  the mother plays a football fanatic and the father plays a quiet man with odd habits . \\
\textbf{(12)}  gallo and ricci arrive at his parent ' s house , and \\
\textbf{(13)}  some extremely funny scenes take place within the house . \\
\textbf{(14)} \textit{ ricci ' s performance during the scene at gallo ' s parent ' s home are very well done . }\\
\textbf{(15)} \textit{ there is constantly humor involved in the interesting dinner table scenes . }\\
\textbf{(16)} \textit{ the way the movie was filmed in this particular part of the movie were interesting and creative . }\\
\textbf{(17)} \textit{ they seemed very mediocre , but they worked out just fine }\\
\textbf{(18)}  . \\
\textbf{(19)} \textit{ gallo ' s character is developed very well . }\\
\textbf{(20)}  the impression that he is very depressed and confused is very clear . \\
\textbf{(21)} \textit{ gallo gives a performance that makes you believe what the character is going through . }\\
\textbf{(22)}  his character goes through many , many problems , just like many people in real life . \\
\textbf{(23)}  this character seemed very realistic to me . \\
\textbf{(24)}  ricci ' s character is funny and different . \\
\textbf{(25)}  she does n ' t care much that she has been kidnaped , in fact , she falls in love the man who kidnaped her ! \\
\textbf{(26)} \textit{\textbf{ ricci is a very wonderful actress and she is starting to get the recognition that she deserves }}\\
\textbf{(27)}  . \\
\textbf{(28)}  buffalo ? \\
\textbf{(29)}  66 is n ' t all laughs though . \\
\textbf{(30)}  many scenes are very dramatic and depressing . \\
\textbf{(31)}  gallo ' s character was so realistic , he was extremely disturbing . \\
\textbf{(32)}  some scenes are supposed to come off as funny , but they actually seemed sad and real to life . \\
\textbf{(33)}  the film sometimes drags along , not giving much material . \\
\textbf{(34)}  i really would have liked to see gallo ' s parents a lot more , and i would have liked to see the characters developed more . \\
\textbf{(35)}  overall , buffalo ? \\
\textbf{(36)}  66 is n ' t as good as some people put it up to be . \\
\textbf{(37)}  the bottom line - \\
\textbf{(38)}  a few hysterical scenes save this film from sinking to the bottom . \\
    \bottomrule
    \end{tabularx}
    \caption{Example of Movies (dev) with gold label \texttt{POS} and predicted label \texttt{NEG}. \textit{Italic} sentences are gold (sentence--level) rationales, \textbf{bold} is the selected rationale.}
    \label{tab:example-appendix-2}
\end{figure*}{}

\end{document}